\documentclass[5p, twocolumn]{elsarticle}[draft]
\usepackage{amssymb}
\usepackage{amsmath}
\usepackage{diagbox}
\biboptions{numbers,sort&compress}
\usepackage[colorlinks]{hyperref}
\usepackage{caption}
\usepackage[below]{placeins}
\begin{document}
\begin{frontmatter}


\title{Reconstruction from edge image combined with color and gradient difference for industrial surface anomaly detection}

\author[Address1]{Tongkun Liu}
\author[Address1,Address2]{Bing Li}
\author[Address1]{Zhuo Zhao\corref{cor1}}
\author[Address1]{Xiao Du}
\author[Address1]{Bingke Jiang}
\author[Address1]{Leqi Geng}
\cortext[cor1]{Corresponding author at: School of Mechanical Engineering, Xi’an Jiaotong University, Xi'an, Shaanxi, China.}

\address[Address1]{State Key Laboratory for Manufacturing System Engineering, Xi’an Jiaotong University,No.99 Yanxiang Road, Yanta District, 710054, Xi’an, Shaanxi, China}%Department and Organization

\address[Address2]{International Joint Research Laboratory for Micro/Nano Manufacturing and Measurement Technologies, Xi’an Jiaotong University,No.99 Yanxiang Road, Yanta District, 710054, Xi’an, Shaanxi, China}%Department and Organization

\begin{abstract}
Reconstruction-based methods are widely explored in industrial visual anomaly detection. Such methods commonly require the model to well reconstruct the normal patterns but fail in the anomalies, and thus the anomalies can be detected by evaluating the reconstruction errors. However, in practice, it's usually difficult to control the generalization boundary of the model. The model with an overly strong generalization capability can even well reconstruct the abnormal regions, making them less distinguishable, while the model with a poor generalization capability can not reconstruct those changeable high-frequency components in the normal regions, which ultimately leads to false positives. To tackle the above issue, we propose a new reconstruction network where we reconstruct the original RGB image from its gray value edges (EdgRec). Specifically, this is achieved by an UNet-type denoising autoencoder with skip connections. The input edge and skip connections can well preserve the high-frequency information in the original image. Meanwhile, the proposed restoration task can force the network to memorize the normal low-frequency and color information. Besides, the denoising design can prevent the model from directly copying the original high-frequent components. To evaluate the anomalies, we further propose a new interpretable hand-crafted evaluation function that considers both the color and gradient differences. Our method achieves competitive results on the challenging benchmark MVTec AD (97.8\% for detection and 97.7\% for localization, AUROC). In addition, we conduct experiments on the MVTec 3D-AD dataset and show convincing results using RGB images only. Our code will be available at https://github.com/liutongkun/EdgRec.
\end{abstract}


\end{frontmatter}


\section{Introduction}
\label{}
Industrial surface anomaly detection aims at identifying all types of visible defects that possibly occur during manufacturing. It plays an important role in manufacturing quality control and has received more and more attention in recent years. Although existing supervised models have achieved good performance in many vision tasks, they are not widely adopted in industrial surface defect detection. The main reason is that a qualified production line can not produce so many defective samples to train such supervised models. More importantly, supervised models are likely to fail on those unseen types of defects that are not included in the training set. However, the above issues can be well mitigated by unsupervised anomaly detection models. For one thing, there are always adequate normal samples in the production line, which are sufficient to train an anomaly detection model without any defective samples. For another, the model trained on normal samples only captures the distribution of the normal data. Therefore, ideally, it's able to detect any unknown defects whose distributions deviate from the normal distribution.

Existing methods for industrial visual anomaly detection can be mainly divided into feature-based methods and reconstruction-based methods. Feature-based methods project the original image into a more distinguishable feature space through ImageNet pre-trained networks or self-supervised tasks. Generally, these methods achieve higher performance than reconstruction-based methods while they are less interpretable and adjustable since it's difficult for engineers to read from those abstract feature vectors. Reconstruction-based methods assume that the model trained on normal samples can only well reconstruct the normal patterns but fail in anomalies, and therefore the anomalies can be detected by comparing the original and reconstructed images through the anomaly evaluation function. Compared to feature-based methods, reconstruction-based methods are easier to understand visually because one can directly observe the differences between the original and reconstructed image. Through designing the specific comparing functions, such methods can be easily adjusted for specific situations with human prior knowledge (e.g., if we only need to detect color anomalies, then we can just design the color comparing function and ignore other differences).

Currently, most reconstruction-based methods are underperforming since it's hard to control the boundary of the model's generalization capability. Concretely, a model which can not reconstruct the anomalies is also likely to fail in reconstructing those variable normal patterns, and vice versa, an overly generalizable model may generalize to those anomalies, i.e., reconstruct the anomalies well and thus makes them less distinguishable. Besides, the image will inevitably be degraded (e.g., generating blurred results in variable regions) during the reconstruction process, which bring challenges to the design of anomaly evaluation function. Directly using pixel-level $l_2$ distance to compare the original and reconstructed image is usually unfavorable since it may cause many false alarms in those degraded normal regions.

\begin{figure}
    \centering
		\includegraphics[scale=.35]{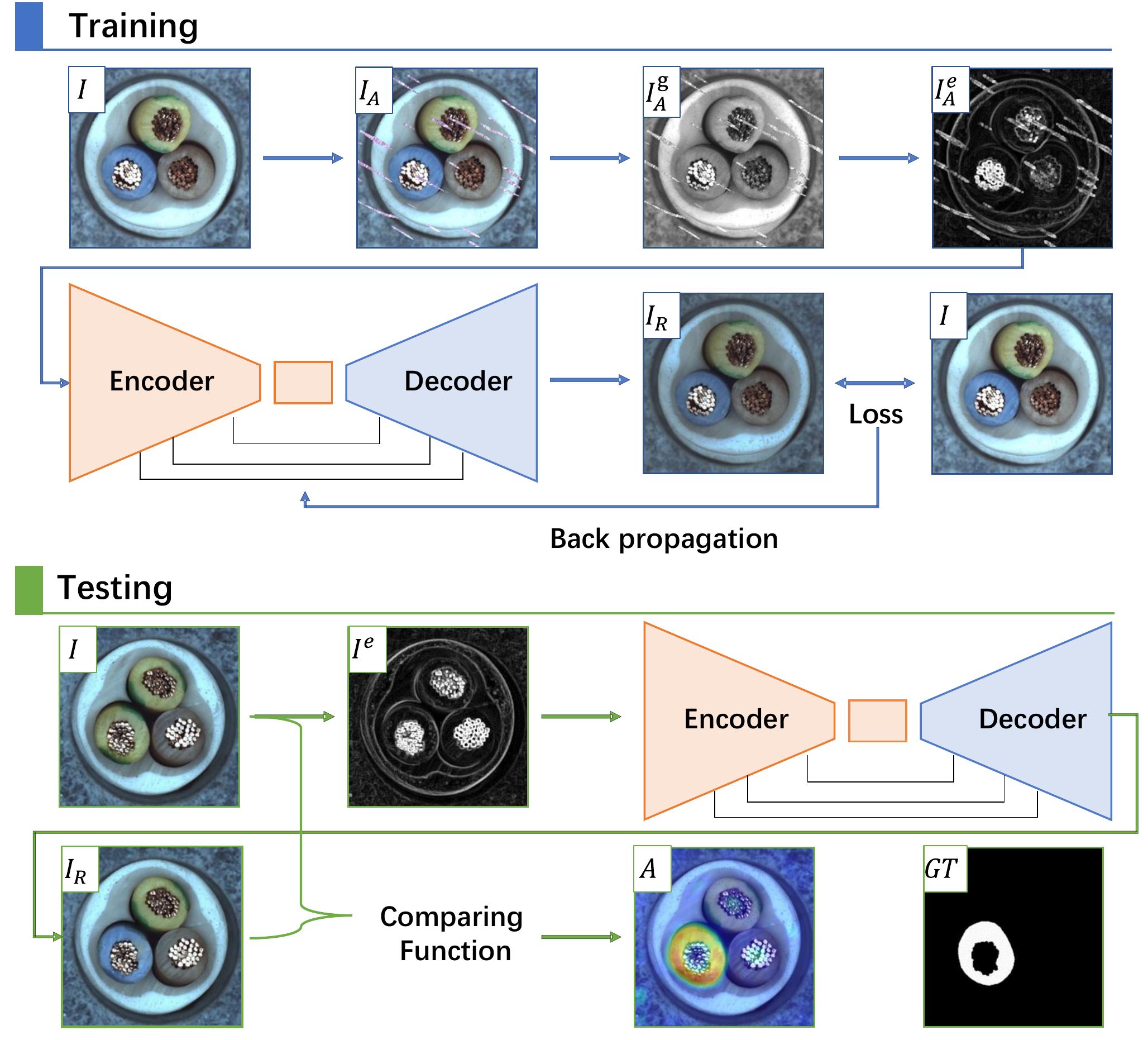}
	\caption{For training phase, we first corrupt the original image $I$ with certain noise and thus get $I_A$; then we convert it to grayscale image $I_A^g$ and extract the edge $I_A^{e}$. Our training goal is to make the reconstructed image $I_R$ as close as possible to the $I$. For testing phase, we extract the grayscale edge $I^e$ of the original test image $I$ and reconstruct it to the RGB image $I_R$. The anomaly map $A$ is obtained by comparing the original and the reconstructed images via the compare function.}
	\label{FIG:1}
\end{figure}

Considering the above issues, this paper propose to boost the performance of the reconstruction-based methods from two aspects. First, we propose to reconstruct the image from its gray value edge, which is motivated by \cite{fei2020attribute}. Since the edge retains only partial contents of the original image, the network needs to generate the removed normal low-frequency and color contents during training. When testing the abnormal images, the model is less likely to generate accurate abnormal patterns as it only sees partial contents (the abnormal edge) of the image. Meanwhile, the input edge preserves the important original high-frequency components, which are usually the hardest parts to be reconstructed \cite{jiang2021focal}. We further use skip connections to reduce the loss of those components in the down-sampling process. Consequently, the edge and skip connections can help better reconstruct those complex high-frequency normal regions and therefore yield fewer false alarms. On the other hand, the above operations may cause the model directly copy the edge from the input. Therefore, we use a denoising autoencoder design to corrupt the original edge with multi scales pseudo anomalies to avoid an identity mapping on the edge region. Fig.\ref{FIG:1} illustrates the main structure of our reconstruction network.

Second, we propose a new color-based evaluation function and combined it with the existing gradient-based function\cite{xue2013gradient,zavrtanik2021reconstruction} as our anomaly evaluation function. This function can effectively detect anomalies while reducing false alarms caused by image degradation in normal areas.
Finally, our contributions are summarized as follows:

(1) We propose a new reconstruction method for industrial visual anomaly detection where we reconstruct the images from their edges. Our specific design can effectively control the generalization capability of the model between anomalies and normal regions

(2) We propose a new color-based anomaly evaluation function to detect color anomalies for reconstruction-based methods. Our function can effectively detect color anomalies and is insensitive to light changes.

(3) We achieve comparable results on the challenging benchmark MVTec AD and the MVTec 3D-AD dataset (using RGB images only) for both anomaly detection and segmentation. Specifically, our anomaly evaluation function is totally hand-crafted and therefore more interpretable and adjustable compared to those latent feature-based evaluation functions (e.g., the separate discriminative network or the perceptual loss).    

\section{Related Work}
\label{20}
Visual anomaly detection is a widely studied topic with applications ranging from medical diagnosis\cite{schlegl2019f}, surveillance\cite{liu2018future}, industrial inspection\cite{tao2022deep}, etc., where there are usually adequate normal samples while abnormal samples are rare and diverse. In this paper, we focus on its application in industrial surface defect detection. This task may be more challenging since it requires the model to not only identify whether there exist anomalies in the image (anomaly detection) but also accurately locate the abnormal areas (anomaly segmentation). Bergmann et al. (2019) \cite{bergmann2019mvtec} propose the MVTec AD (detailed in Sec \ref{4.1}), a comprehensive dataset for industrial visual anomaly detection including 15 different industrial products. This dataset quickly became the most convincing benchmark in industrial visual anomaly detection and has sparked much research. Here we mainly divided the existing research into two groups: feature-based and reconstruction-based.

\subsection{Feature based methods}
\label{2.1}
Feature-based methods aim to find a feature space where the normal and abnormal features are fully distinguishable. Since there exists no abnormal sample during training, it's preferable to leverage the ImageNet pre-trained network\cite{bergmann2020uninformed, roth2021towards,yu2021fastflow,defard2021padim,zheng2021focus,rippel2021modeling,shi2021unsupervised} or the models obtained through self-supervised tasks\cite{li2021cutpaste,yi2020patch,sohn2020learning}, as feature extractors. For the pre-trained network, several studies\cite{roth2021towards,shi2021unsupervised} have found that it's important to select appropriate hierarchy levels of features, because the low-level features lack global awareness, while the extremely high-level features may be biased to the pre-trained task itself. Also, the pre-trained network can be used as a teacher network to detect anomalies by knowledge distillation\cite{bergmann2020uninformed}. For the self-supervised based methods, the key is to design suitable auxiliary tasks. Li et al.\cite{li2021cutpaste} propose to use Cutpaste augmentation to train a one-class classifier. Other auxiliary tasks includes the position prediction\cite{yi2020patch}, the geometric transformation prediction\cite{sohn2020learning}, etc.  

Overall, benefiting from the powerful representation capabilities of deep features, feature-based methods can achieve better performance compared to existing reconstruction-based methods. In particular, \cite{roth2021towards} achieves state-of-the-art performance on the MVTec AD. However, these methods are hard to be optimized for the specific situation since those deep features are too abstract to introduce prior knowledge.

\subsection{Reconstruction based methods}
\label{2.2}
Reconstruction-based methods commonly leverage generative models such as autoencoders\cite{bergmann2018improving,collin2021improved,zavrtanik2021reconstruction}, VAEs\cite{venkataramanan2020attention}, GANs\cite{schlegl2019f,liang2022omni}, etc., to detect anomalies in the image space. Generally, these methods contain two steps: 1. Reconstruct the image; 2. Compare the original and reconstructed images to get anomaly maps. 

 \textbf{Reconstruct the image}. Early works mainly leverage denoising autoencoders\cite{mei2018unsupervised, kang2018deep,collin2021improved} to help the network better capture the normal distribution and avoid learning an identity mapping. In the training phase, these methods corrupt the original image with certain noise and make the network eliminate it. In addition to some low-level noise like Gaussian noise, cutout, stain, etc., the image can also be corrupted by some semantic transformations, like the geometric transformation\cite{golan2018deep,hendrycks2019using,huang2019inverse}, color transformation\cite{fei2020attribute}, inpainting masks\cite{haselmann2018anomaly,zavrtanik2021reconstruction,pirnay2022inpainting} etc., which are summarized into an attribute removal-and-restoration framework by Ye et al. \cite{fei2020attribute}. They argue that the network can learn more robust features during the process of restoring the previously removed attributes. Following this paradigm, we propose a specific attribute removal-restoration task where the low-frequency and color attributes are the main attributes to be restored.
 
\textbf{Compare the images}. After the reconstruction, the anomalies can be detected by comparing the original and reconstructed images. Early comparing functions include $l_2$ distance, structure similarity (SSIM) \cite{wang2004image}, etc. Furthermore, Zavrtanik et al. \cite{zavrtanik2021reconstruction} introduce a multi-scale gradient map (MSGMS) anomaly evaluation function which significantly boosts the performance. However, MSGMS performs poorly on those low-frequency color anomalies. Later, Zavrtanik et al. \cite{zavrtanik2021draem} further propose to use a separate discriminative network (DRAEM) which takes the concatenation of the original and reconstructed images as input and detects the anomalies via image segmentation. While DRAEM achieves remarkable performance on the MVTec AD, the additional discriminative network introduces extra latent features and therefore makes the segmentation results less interpretable. Similarly, the current state-of-the-art reconstruction-based method OCR-GAN \cite{liang2022omni} also leverage latent space features and combine them with $l_1$ distance to detect anomalies. Differently, in this paper, we still focus on hand-crafted anomaly score functions, which are more interpretable and adjustable. Concretely, we propose a new color comparing function and combine it with the existing MSGMS function. The proposed function can effectively detect various anomalies. 

\section{Methods}
\label{30}
Our reconstruction framework is based on an UNet-type encoder-decoder network with the corrupted grayscale edge as input. Specifically, we first corrupted the original image with certain noises; then we convert the corrupted image into a grayscale edge; after that, we train a network to reconstruct the original image from its corrupted edge; finally, we discuss how to design the anomaly evaluation function.

\subsection{Get the corrupted edges}
\label{3.1}
Our basic idea is to formulate an attribute removal-and-restoration task that can be suitable for various industrial anomaly detection scenarios. Specifically, we construct a `grayscale edge to RGB image' task where we remove the low frequency and color attributes in the original image and train a network to restore them. This design is based on two considerations. First, low-frequency and color contents are general attributes in various images. We notice that there also exist other tasks such as the restoration of the geometrically transformed image \cite{sohn2020learning,golan2018deep,hendrycks2019using, huang2019inverse}, However, compared with our design, these methods are less general, e.g., the above geometric transformation framework cannot be applied to spatially invariant textures, while our design can be applied to both texture and object images. Second, preserving edge information enables the network to better reconstruct the details in normal patterns, which can effectively reduce the false positive rate in complex normal areas. On the other hand, preserving the edges may also lead to the model producing identity mappings of the original high-frequency components. To avoid this, we first corrupt the original image with certain noise.

\begin{figure}
    \centering
		\includegraphics[scale=.25]{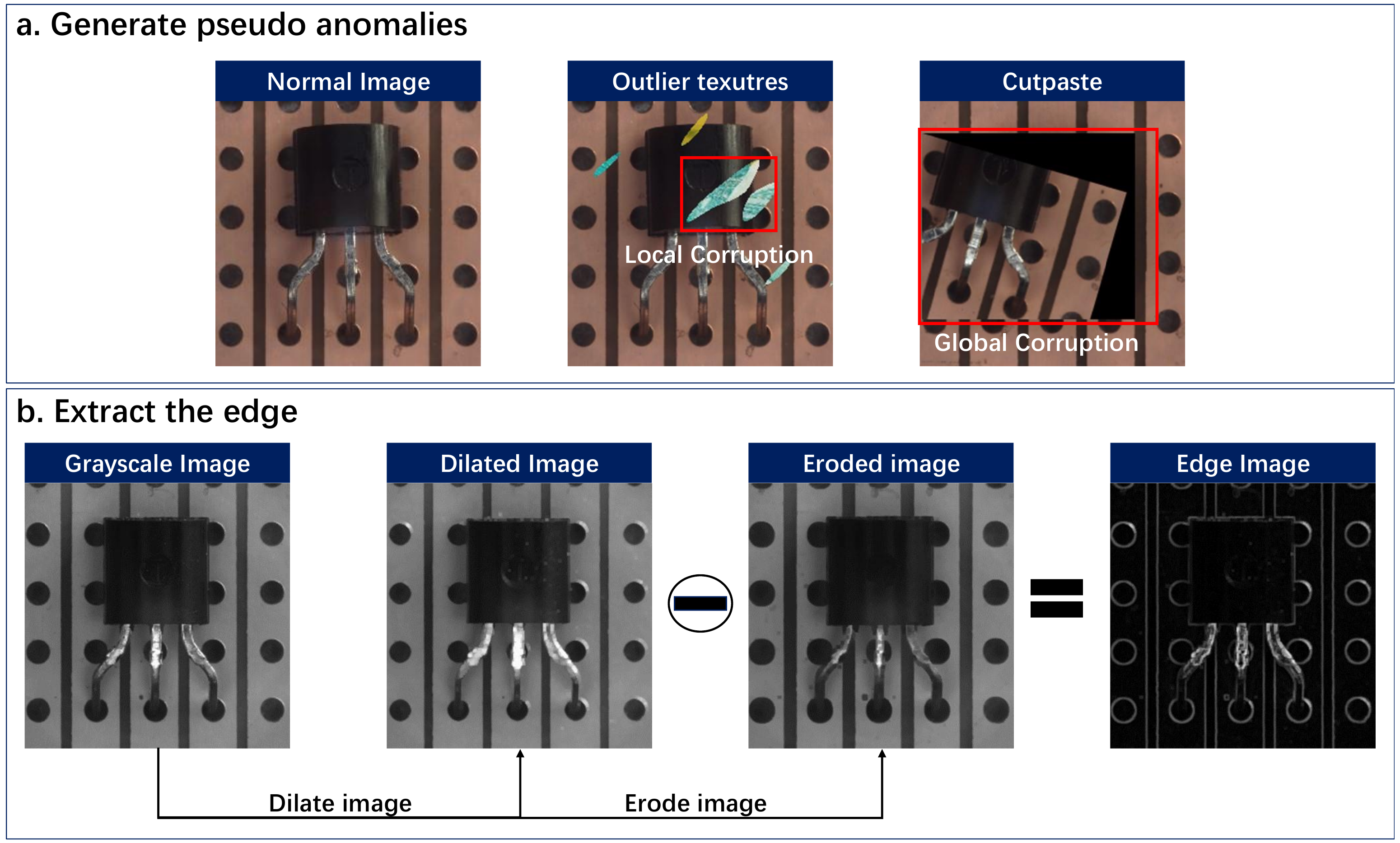}
	\caption{(a). Generate pseudo anomalies. We consider both the local (from the DTD dataset) and global (by Cutpaste augmentations) corruption when generating pseudo anomalies. (b). We use morphological erosion and dilation operations to extract the grayscale edge.}
	\label{FIG:2}
\end{figure}

We adopt the strategy proposed in \cite{zavrtanik2021draem} to generate simulated anomalies whose textures are from external texture dataset DTD \cite{cimpoi2014describing} with the shapes of randomly generated Perlin noise. However, we observe that if only use these out-of-distribution textures as pseudo anomalies, the model cannot well distinguish the foreground and background areas. This makes it difficult to detect structural defects caused by missing components. Therefore, we additionally introduce large-scale Cutpaste augmentations \cite{li2021cutpaste}, which make the model pay more attention to the global information, as shown in Fig. \ref{FIG:2}, a. Finally, we use the above-mentioned multi-scale pseudo anomalies to corrupted the original image and activate each corruption with certain probabilities. The corrupted image is then converted to grayscale. To obtain its edge, we use a $3 \times 3$ convolution kernel to dilate and erode the original image separately and make a difference, as shown in Fig. \ref{FIG:2}, b.  

\subsection{Reconstruction network and loss function}
\label{3.2}
\begin{figure}
    \centering
		\includegraphics[scale=.3]{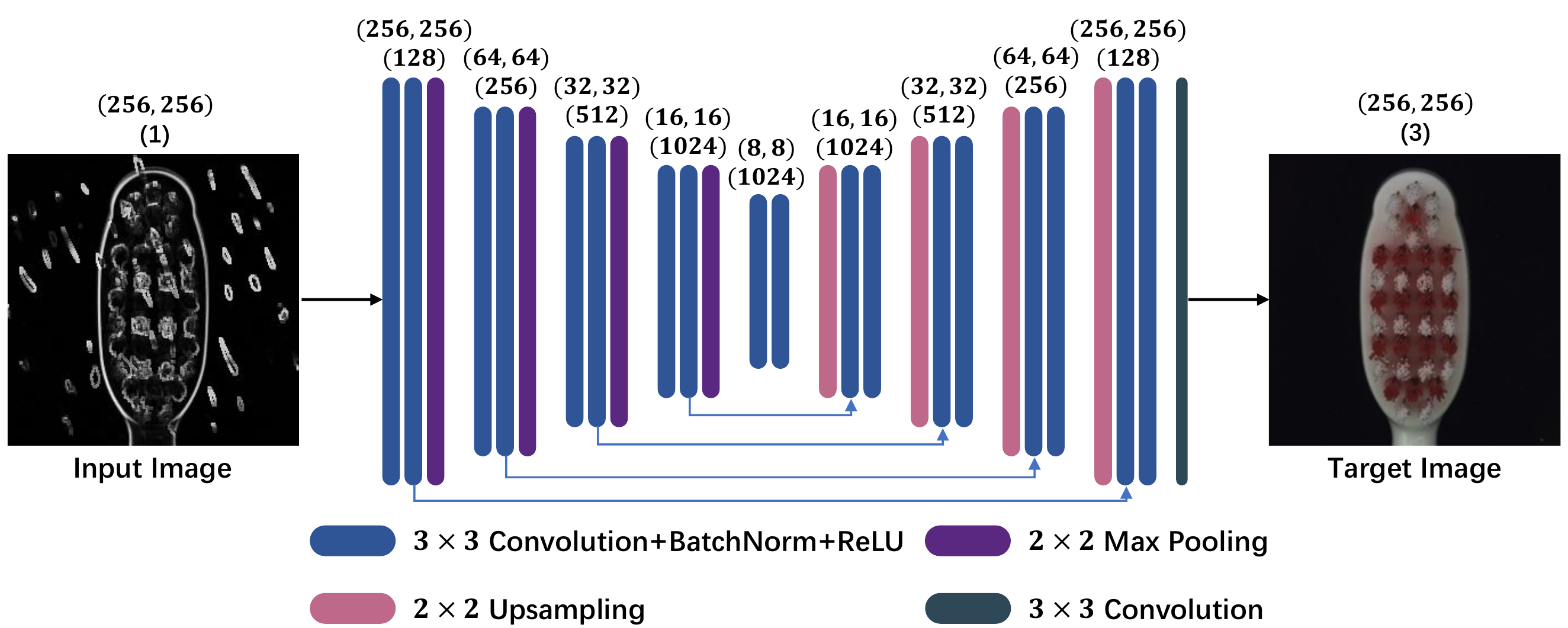}
	\caption{The architecture of the proposed reconstruction network.}
	\label{FIG:3}
\end{figure}

The architecture of the proposed network is shown in Fig. \ref{FIG:3}. Specifically, we add skip connections between the corresponding encoding and decoding layers. For many reconstruction networks \cite{zavrtanik2021draem,zavrtanik2021reconstruction}, adding skip connections is not preferred for anomaly detection since those shallow encoded features can be directly transferred to the final decoding layers, which increases the risk of the model generating an identity map. \cite{hou2021divide} even particularly designs a memory module to filter the features in skip connections. In our work, the encoded features (from the grayscale edge) are quite different from the content to be decoded (the RGB image). Consequently, it's less likely that our model uses skip connections to take `shortcuts'. Experimentally, we find that using skip connections can boost our final performance because it can reduce the reconstruction errors of normal areas.

For the loss function, we adopt a common loss setting in image reconstruction, that is, a pixel-based $l_2$ loss and a patch-based SSIM loss \cite{bergmann2018improving}. Among them, SSIM loss enables the model to better capture the structured information and thus reduces the reconstruction blurring. Finally, our loss function is defined as: \begin{equation}
L(I,I_r)= l_2(I,I_r)+\lambda L_{\rm SSIM} (I,I_r)
\label{1}
\end{equation}
where $I$ and $I_r$ are the original and reconstructed image respectively, and $\lambda$ is an adjustable hyperparameter. For all the experiments in this paper, we set $\lambda$ as 1.

\subsection{Anomaly Evaluation function}
\label{3.3}
We design a hand-craft anomaly evaluate function to compare the original and reconstructed image. Specifically, our function considers both the color and gradient difference to detect anomalies.

\textbf{Color difference}. To evaluate the color difference, we first convert the original and reconstructed image from RGB color space to CIELAB color space. CIELAB color space is defined by the International Commission on Illumination in 1976. It is more consistent with human perception and consists of three components: \textbf{L} for perceptually lightness, and \textbf{a} and \textbf{b} for the unique colors of human vision. The process of converting an RGB image to a CIELAB image can be written as:
\begin{equation}
\left(\begin{array}{l}
X \\
Y \\
Z
\end{array}\right)=\left(\begin{array}{lll}
0.412453 & 0.357580 & 0.180423 \\
0.212671 & 0.715160 & 0.072169 \\
0.019334 & 0.119193 & 0.950227
\end{array}\right)\left(\begin{array}{c}
R \\
G \\
B
\end{array}\right)
\label{2}
\end{equation}

\begin{equation}
\begin{aligned}
L &= 116 f\left(\frac{Y}{Y_{w}}\right)-16 \\
a &= 500\left(f\left(\frac{X}{X_{w}}\right)-f\left(\frac{Y}{Y_{w}}\right)\right) \\
b &= 200\left(f\left(\frac{Y}{Y_{w}}\right)-f\left(\frac{Z}{Z_{w}}\right)\right)
\end{aligned}
\label{3}
\end{equation}
where

\begin{equation}
f(t)=\left\{\begin{array}{ll}
t^{\frac{1}{3}}, & \text { if } \space \space t> (\frac{6}{29})^3 \\
\left(\frac{841}{108}\right) t +\frac{4}{29} , & \text { else }
\end{array}\right.
\label{4}
\end{equation}and $X, Y, Z$ represent the intermediate values from the CIEXYZ space during the conversion.  $X_{w}, Y_{w}, Z_{w}$ describe a specified white achromatic reference illuminant. For our experiments, we use the default settings in OpenCV Library.

We find that in the reconstruction task, the degradation of the normal area is mainly reflected in \textbf{L}. We guess that this is caused by the instability of the lighting during the acquisition of the image, which makes the model tend to take an average of all light intensities. However, \textbf{a} and \textbf{b} are less sensitive to changes in light intensity. Therefore, for the original image $I$ and the reconstructed image $I_r$, we design our color evaluation function as:
\begin{equation}
A_{color}=[(I_r^{a}-I^{a})^2+(I_r^{b}-I^{b})^2]*f_{mean}
\label{50}
\end{equation}
where we use an additional mean filter $f_{mean}$ with a kernel size of 11 to smooth the final result and $*$ refers to the convolution operation.

\textbf{Structure difference}. To evaluate the structure difference, we leverage the multi-scale gradient magnitude similarity (MSGMS) proposed in \cite{zavrtanik2021reconstruction}. Concretely, MSGMS calculates the gradient magnitude similarities (GMS) \cite{xue2013gradient} between the original and reconstructed image at different scales through the image pyramid and takes the average value. The structure difference evaluation function can be written as:
\begin{equation}
A_{strucutre}=1_{H\times W}-{\rm MSGMS} (I,I_r)* f_{mean}
\label{5}
\end{equation}
where $1_{H\times W}$ refers to an identity matrix with the same height $H$ and width $W$ as the original image. Here we set the kernel size of $f_{mean}$ as 21.

Finally, we combine $A_{color}$ and $A_{strucutre}$ as our pixel-level anomaly evaluation function $A_{score}$. Specifically, we introduce a scaling factor $c$ to make $A_{color}$ and $A_{strucutre}$ in the same order of magnitude, since they are calculated with different data types ( $A_{color}$ is calculated in uint8, that is, [0-255], while $A_{strucutre}$ is calculated in [0-1] ). For all the experiments in this paper, the $c$ is set as $1e^{-3}$. For the image-level detection score, we choose the maximum value of the anomaly map.
\begin{equation}
A_{score}=cA_{color} + A_{strucutre}
\label{6}
\end{equation}

\begin{table*}[]
\centering
\footnotesize
\caption{Comparison of image-level detection results on the MVTec AD dataset. (AUROC\%)}
\label{Table1}
\scalebox{0.8}{
\begin{tabular}{c|ccccccccccc}
\hline
Category   & AE-$l_2$ \cite{bergmann2018improving} & AE-SSIM \cite{bergmann2018improving} & Patch-SVDD \cite{yi2020patch}   & NSA \cite{schluter2021self}          & Diff \cite{rudolph2021same}          & ITAE \cite{huang2019inverse}         & RIAD \cite{zavrtanik2021reconstruction}          & DFR \cite{shi2021unsupervised}         & CutPaste \cite{li2021cutpaste}       & Intra \cite{pirnay2022inpainting}          & EdgRec         \\ \hline
Carpet     & 59    & 87      & 92.9         & 95.6         & 92.9          & 70.6         & 84.2           & 97          & \textbf{100.0} & 98.8           & 97.4           \\
Grid       & 90    & 94      & 94.7         & 99.9         & 84.0          & 88.3         & 99.6           & 98          & 99.1           & \textbf{100.0} & 99.7           \\
Leather    & 75    & 78      & 90.9         & 99.9         & 97.1          & 86.2         & \textbf{100.0} & 99          & \textbf{100}   & \textbf{100.0} & \textbf{100.0} \\
Tile       & 51    & 59      & 97.8         & \textbf{100} & 99.4          & 73.5         & 93.4           & 86          & 99.8           & 98.2           & \textbf{100.0} \\
Wood       & 73    & 73      & 96.5         & 97.5         & \textbf{99.8} & 92.3         & 93.0           & 94          & \textbf{99.8}  & 98.0           & 94.0           \\ \hline
Avg.tex    & 69.6  & 78.2    & 94.5         & 98.6         & 94.6          & 82.2         & 95.1           & 94.8        & \textbf{99.7}  & 99.0           & 98.2           \\ \hline
Bottle     & 86    & 93      & 98.6         & 97.7         & 99.0          & 94.1         & 99.9           & 97          & \textbf{100}   & \textbf{100.0} & \textbf{100.0} \\
Cable      & 86    & 82      & 90.3         & 94.5         & 95.9          & 83.2         & 81.9           & 92          & 96.2           & 84.2           & \textbf{97.9}  \\
Capsule    & 88    & 94      & 76.7         & 95.2         & 86.9          & 68.1         & 88.4           & \textbf{99} & 95.4           & 86.5           & 95.5           \\
Hazelnut   & 95    & 97      & 92.0         & 94.7         & 99.3          & 85.5         & 83.3           & \textbf{99} & 99.9           & 95.7           & 98.4           \\
Metal Nut  & 86    & 89      & 94.0         & 98.7         & 96.1          & 66.7         & 88.5           & 93          & \textbf{98.6}  & 96.9           & 97.3           \\
Pill       & 85    & 91      & 86.1         & 99.2         & 88.8          & 78.6         & 83.8           & 97          & 93.3           & 90.2           & \textbf{99.0}  \\
Screw      & 96    & 96      & 81.3         & 90.2         & 96.3          & \textbf{100} & 84.5           & 99          & 86.6           & 95.7           & 89.9           \\
Toothbrush & 93    & 82      & \textbf{100} & 100.0        & 98.6          & \textbf{100} & \textbf{100}   & 99          & 90.7           & 99.7           & \textbf{100.0} \\
Transistor & 86    & 90      & 91.5         & 95.1         & 91.1          & 84.3         & 90.9           & 80          & 97.5           & 95.8           & \textbf{99.8}  \\
Zipper     & 77    & 88      & 97.9         & 99.8         & 95.1          & 87.6         & 98.1           & 96          & \textbf{99.9}  & 99.4           & 98.3           \\ \hline
Avg.obj    & 87.8  & 90.2    & 90.8         & 96.5         & 94.7          & 84.8         & 89.9           & 95.1        & 95.8           & 94.4           & \textbf{97.6}  \\ \hline
Avg.all    & 82    & 86      & 92.1         & 97.2         & 94.7          & 83.9         & 91.7           & 95          & 97.1           & 95.9           & \textbf{97.8}  \\ \hline
\end{tabular}}
\end{table*}

\begin{table}[]
\centering
\footnotesize
\caption{Comparison of pixel-level localization results on the MVTec AD dataset. (AUROC\%/AP\%)}
\label{Table2}
\scalebox{0.85}{
\begin{tabular}{c|cccc}
\hline
Category   & ST \cite{bergmann2020uninformed}        & RIAD \cite{zavrtanik2021reconstruction}               & PaDim \cite{defard2021padim}              & EdgRec             \\ \hline
Carpet     & 93.5/52.2 & 96.3/61.4          & 99.0/60.7          & \textbf{99.4/79.8} \\
Grid       & 89.9/10.1 & 98.8/36.4          & 97.1/35.7          & \textbf{99.2/46.6} \\
Leather    & 97.8/40.9 & 99.4/49.1          & 99.0/53.5          & \textbf{99.7/66.4} \\
Tile       & 92.5/65.3 & 89.1/52.6          & 94.1/52.4          & \textbf{98.6/84.5} \\
Wood       & 92.1/53.3 & 85.8/38.2          & \textbf{94.1/46.3} & \textbf{91.4/54.8} \\ \hline
Avg.tex    & 93.2/44.4 & 93.9/47.5          & 96.7/49.7          & \textbf{97.7/66.4} \\ \hline
Bottle     & 97.8/74.2 & \textbf{98.4/76.4} & 98.2/77.3          & \textbf{98.3/77.9} \\
Cable      & 91.9/48.2 & 84.2/24.4          & 96.7/45.4          & \textbf{97.7/70.8} \\
Capsule    & 96.8/25.9 & 92.8/38.2          & \textbf{98.6/46.7} & 95.2/40.5          \\
Hazelnut   & 98.2/57.8 & 96.1/33.8          & 98.1/61.1          & \textbf{99.4/81.7} \\
Metal Nut  & 97.2/83.5 & 92.5/64.3          & 97.3/77.4          & \textbf{98.0/79.9} \\
Pill       & 96.5/62.0 & 95.7/51.6          & 95.7/61.2          & \textbf{98.7/79.3} \\
Screw      & 97.4/7.8  & \textbf{98.8/43.9} & 98.4/21.7          & 97.7/42.9          \\
Toothbrush & 97.9/37.7 & 98.9/50.6          & 98.8/54.7          & \textbf{99.2/59.2} \\
Transistor & 73.7/27.1 & 87.7/39.2          & \textbf{97.6/72.0} & 94.3/71.2          \\
Zipper     & 95.6/36.1 & 97.8/63.4          & \textbf{98.4/58.2} & \textbf{98.7/52.5} \\ \hline
Avg.obj    & 94.3/46.0 & 94.4/48.5          & 97.8/57.6          & \textbf{97.7/65.5} \\ \hline
Avg.all    & 93.9/45.5 & 94.2/48.2          & 97.4/55.0          & \textbf{97.7/65.8} \\ \hline
\end{tabular}}
\end{table}

\begin{table}[]
\centering
\footnotesize
\caption{Comparison of pixel-level localization results on the MVTec AD dataset. (AUPRO\%)}
\label{Table3}
\scalebox{0.85}{
\begin{tabular}{c|ccccc}
\hline
Category   & AE-SSIM \cite{bergmann2018improving} & Ano-GAN \cite{schlegl2019f} & ST \cite{bergmann2020uninformed}           & NSA \cite{schluter2021self}           & EdgRec        \\ \hline
Carpet     & 65      & 20      & 87.9          & 85.0          & \textbf{96.9} \\
Grid       & 85      & 23      & 95.2          & 96.8          & \textbf{97.0} \\
Leather    & 56      & 38      & 94.5          & 98.7          & \textbf{98.8} \\
Tile       & 18      & 18      & 94.6          & 95.3          & \textbf{96.3} \\
Wood       & 61      & 39      & \textbf{91.1} & 85.3          & 77.5          \\ \hline
Avg.tex    & 57      & 27.6    & 92.7          & 92.2          & \textbf{93.3} \\ \hline
Bottle     & 83      & 62      & 93.1          & 92.9          & \textbf{94.3} \\
Cable      & 48      & 38      & 81.8          & \textbf{89.9} & 88.7          \\
Capsule    & 86      & 31      & \textbf{96.8} & 91.4          & 82.2          \\
Hazelnut   & 92      & 70      & \textbf{96.5} & 93.6          & 95.4          \\
Metal Nut  & 60      & 32      & 94.2          & \textbf{94.6} & 91.2          \\
Pill       & 83      & 78      & \textbf{96.1} & 96.0          & \textbf{96.1} \\
Screw      & 89      & 47      & \textbf{94.2} & 90.1          & 89.3          \\
Toothbrush & 78      & 75      & 93.3          & 90.7          & \textbf{94.9} \\
Transistor & 73      & 55      & 66.6          & 75.3          & \textbf{87.7} \\
Zipper     & 67      & 47      & 95.1          & 89.2          & \textbf{96.0} \\ \hline
Avg.obj    & 75      & 53.7    & 90.8          & 90.4          & \textbf{91.5} \\ \hline
Avg.all    & 69      & 45      & 91.4          & 91.0          & \textbf{92.1} \\ \hline
\end{tabular}}
\end{table}

\begin{table}[]
\footnotesize
\centering
\caption{Comparison of image-level detection results with reconstruction-based methods on the MVTec AD dataset. (AUROC\%)}
\label{Table4}
\begin{tabular}{c|cc}
\hline
Methods & performance & anomaly evaluation function \\ \hline
RIAD \cite{zavrtanik2021reconstruction}    & 91.7        & hand-crafted             \\
InTra \cite{pirnay2022inpainting}   & 95.9        & hand-crafted             \\
DRAEM \cite{zavrtanik2021draem}   & 98.0        & discriminative network   \\
OCR-GAN \cite{liang2022omni} & 98.3        & $l_1$ and latent-feature    \\
EdgRec  & 97.8        & hand-crafted             \\ \hline
\end{tabular}
\end{table}

\begin{table*}[]
\footnotesize
\caption{Comparison of image-level detection results on the MVTec 3D-AD dataset.(AUROC\%). The results of baselines are from \cite{zheng2022benchmarking}}
\label{Table5}
\centering
\begin{tabular}{c|ccccccc}
\hline
Category    & PaDim \cite{defard2021padim}         & PatchCore \cite{roth2021towards}     & FastFlow \cite{yu2021fastflow}      & CFlow \cite{gudovskiy2022cflow} & Diff \cite{rudolph2021same}  & CSflow \cite{rudolph2022fully}        & EdgRec      \\ \hline
bagle       & \textbf{97.5} & 91.2          & 89.3          & 88.0  & 81.9 & 89.4          & 89.7          \\
cable gland & 77.5          & 90.2          & 62.0          & 85.8  & 67.0 & \textbf{91.7} & 85.5          \\
carrot      & 69.8          & \textbf{88.5} & 79.5          & 82.8  & 61.2 & 74.9          & \textbf{88.5} \\
cookie      & 58.2          & \textbf{70.9} & 42.6          & 56.3  & 48.4 & 66.8          & 70.1          \\
dowel       & \textbf{95.9} & 95.2          & 88.0          & 98.6  & 63.4 & 93.8          & 94.5          \\
foam        & 66.3          & 73.3          & 72.8          & 73.8  & 68.9 & \textbf{89.7} & 82.1          \\
peach       & \textbf{85.8} & 72.7          & 65.1          & 75.7  & 65.5 & 60.3          & 67.9          \\
potato      & 53.5          & 56.2          & 56.0          & 62.8  & 60.0 & 41.9          & \textbf{75.5} \\
rope        & 83.2          & 96.2          & \textbf{98.2} & 97.0  & 72.9 & 97.1          & 96.8          \\
tire        & 76.0          & 76.8          & 61.3          & 72.0  & 53.6 & 72.6          & \textbf{81.8} \\ \hline
average     & 76.4          & 81.1          & 71.5          & 79.3  & 64.3 & 77.8          & \textbf{83.2} \\ \hline
\end{tabular}
\end{table*}

\begin{table}[]
\centering
\footnotesize
\caption{Comparison of pixel-level detection results on the MVTec 3D-AD dataset(AUPRO\%). The results of baselines are from \cite{zheng2022benchmarking}}
\label{Table6}
\scalebox{0.9}{
\begin{tabular}{c|cccc}
\hline
Category    & PatchCore \cite{roth2021towards}    & FastFlow \cite{yu2021fastflow} & CFlow \cite{gudovskiy2022cflow}         & EdgRec      \\ \hline
bagle       & 89.9          & 88.0     & 85.5          & \textbf{94.4} \\
cable gland & 95.3          & 75.2     & 91.9          & \textbf{96.3} \\
carrot      & 95.7          & 92.3     & 95.8          & \textbf{97.6} \\
cookie      & \textbf{91.8} & 81.2     & 86.7          & 76.9          \\
dowel       & 93.0          & 92.9     & \textbf{96.9} & 96.7          \\
foam        & 71.9          & 64.6     & 50.0          & \textbf{87.6} \\
peach       & \textbf{92.0} & 78.2     & 88.9          & 91.0          \\
potato      & 93.7          & 61.5     & 93.5          & \textbf{96.0} \\
rope        & \textbf{93.8} & 91.3     & 90.4          & 93.3          \\
tire        & \textbf{92.9} & 55.0     & 91.9          & 90.7          \\ \hline
average     & 91.0          & 78.0     & 87.1          & \textbf{92.1} \\ \hline
\end{tabular}}
\end{table}

\section{Experiments}
\subsection{Datasets}
\label{4.1}
\textbf{MVTec AD}. The MVTec AD dataset \cite{bergmann2019mvtec} is specifically developed for 2D unsupervised industrial visual anomaly detection and is currently the most important benchmark in this field. The dataset covers 15 different industrial products where there are 5 texture categories and 10 object categories. For the training set, each category consists of 60-320 normal images and 3629 images in total. For the testing set, each category consists of 42-167 normal and abnormal images and 1725 images in total. The anomalies vary in color, shape, and scale, requiring the model to take into account both the global semantic information and local details.

\textbf{MVTec 3D-AD}. The MVTec 3D-AD dataset \cite{bergmann2021mvtec} is also developed for industrial anomaly detection. Differently, this dataset contains both the RGB images and 3D scans acquired by a high-resolution industrial 3D sensor. As our method only considers the 2D detection issue, we exclusively use RGB images for training and testing. The dataset consists of 10 different categories. For each category, there exist 210-361 normal images for training and 101-159 normal and abnormal images for testing. Overall, the training set contains 2625 images, the validation set contains 294 images, and the testing set contains 1197 images. 

\subsection{Evaluation metrics}
We evaluate our approach with different metrics so as to be comparable with other baselines. Basically, we use the area under the curve (AUC) of the receiver operating characteristics (ROC) to evaluate the performances of image-level anomaly detection and pixel-level anomaly localization. Considering many anomalies only occupy a small fraction of pixels in the image, the AUROC may not well reflect the actual anomaly localization performance. Therefore, we additionally introduce the pixel-wise average precision metric (AP), which is also adopted by \cite{zavrtanik2021draem}, and the area under the per-region-overlap curve (AUC-PRO), which is proposed by \cite{bergmann2020uninformed}. Specifically, when using the AUC-PRO, we follow the standard setting where the value is integrated from 0 to 0.3 across FPRs (average per-pixel false positive rate). 

\subsection{Implementation Details}
Our work is implemented in Pytorch with an NVIDIA GeForce GTX 3090Ti. We resize all the original images of the MVTec AD and the MVTec 3D-AD to $256\times 256$ for both training and testing. For each category of these two datasets, we train the model from scratch for 800 epochs with a batch size of 8. We use the Adam optimizer with an initial learning rate of $1e^-4$, and the learning rate is multiplied by 0.2 after 640 and 720 epochs. The weight decay is set to 0 for the MVTec AD and $1e^-3$ for the MVTec 3D-AD. We use image rotation in the range of ($-45^{\circ}, 45^{\circ})$ as data augmentation for all the training tasks.

\subsection{Comparison with baselines}
\begin{figure*}
    \centering
		\includegraphics[scale=0.2]{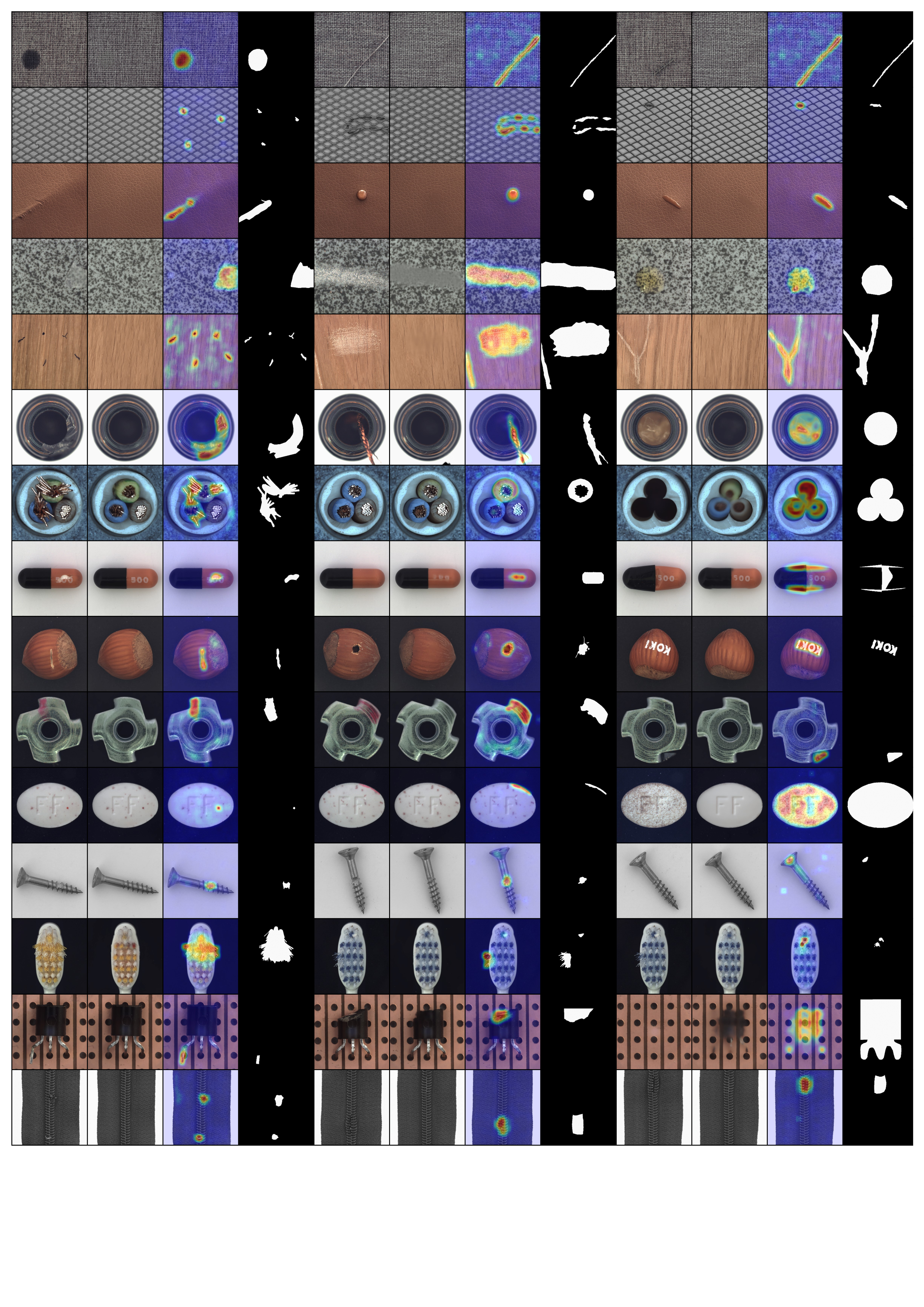}
    	\caption{Qualitative results on the MVTec AD dataset. We show the original image, the reconstructed image, the anomaly map, and the groud truth from left to right.}
	\label{FIG:7}
\end{figure*}

\textbf{MVTec AD}. We compare our performance with the early baselines \cite{bergmann2018improving} and several state-of-the-art methods on the MVTec AD, including both feature-based \cite{yi2020patch, schluter2021self,rudolph2021same,shi2021unsupervised,li2021cutpaste,bergmann2020uninformed,defard2021padim}, and reconstruction-based methods \cite{huang2019inverse,zavrtanik2021reconstruction,pirnay2022inpainting}. Table \ref{Table1} shows the image-level detection result. Table \ref{Table2} and \ref{Table3} show the pixel-level anomaly segmentation results. Our approach outperforms many existing state-of-the-art methods. Specifically, we compare the overall results with existing state-of-the-art reconstruction-based methods in Table \ref{Table4}. We outperform RIAD and Intra, which also use hand-crafted evaluation functions. This can be attributed to our restoration task design and the additional color evaluation function. RIAD and Intra also follow the attribute-removal-restoration paradigm where they leverage image inpainting to build restoration tasks. However, this paradigm may be inefficient since the network can only reconstruct the partial image (the masked part) in a single pass, while our restoration task can directly reconstruct the entire image. On the other hand, our performance is lower than DRAEM \cite{zavrtanik2021draem} and OCR-GAN \cite{liang2022omni}. However, compared with these methods, our main advantages lie in that we all use hand-crafted evaluation functions, which are more interpretable and adjustable in practice. Meanwhile, our method is more efficient than DRAEM as we only use a single reconstruction network. The examples of qualitative anomaly localization results can be seen in Fig. \ref{FIG:7}.

\begin{figure*}
    \centering
		\includegraphics[scale=0.2]{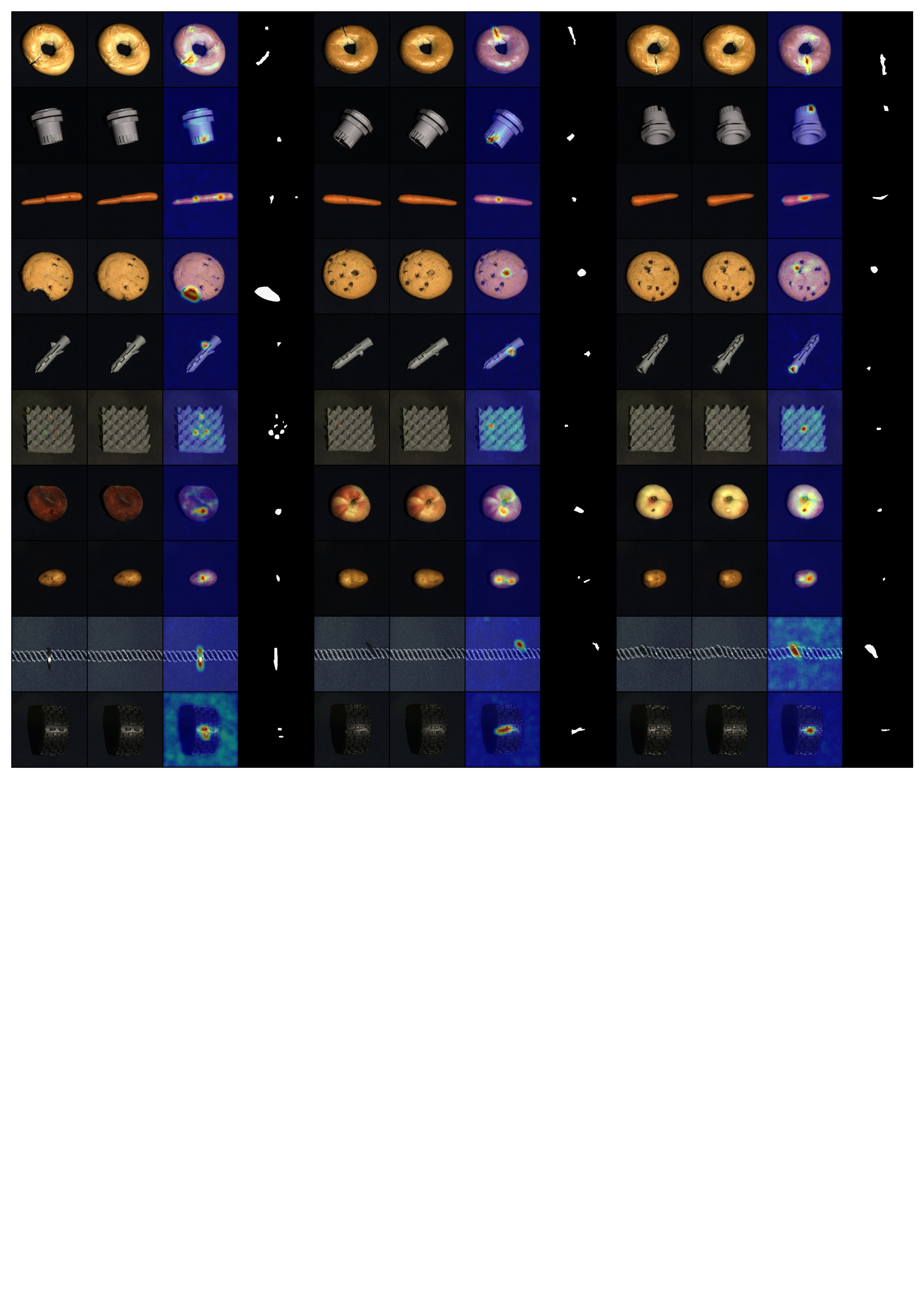}
	\caption{Qualitative results on the MVTec 3D-AD dataset. We show the original image, the reconstructed image, the anomaly map, and the groud truth from left to right.}
	\label{FIG:8}
\end{figure*}

\textbf{MVTec 3D-AD}. The detection and localization performances on the MVTec 3D-AD \cite{bergmann2021mvtec} are shown in Table \ref{Table5} and \ref{Table6} respectively. Overall, the results are still underperforming when only using 2D images since some anomalies are only detectable in 3D space.  However, we still outperform several state-of-the-art 2D visual anomaly detection models.  Fig. \ref{FIG:8} shows some qualitative results. On this dataset, a better strategy should be to combine both 3D and color features. To this end, as our approach achieves good results on RGB images, hopefully, it can be integrated with other point cloud processing algorithms in the future to build more effective multimodal detection algorithms.

 \section{Discussion and ablation studies.}
In this section, we discuss the effectiveness of different components in our method through various ablation studies on the MVTec AD dataset. These components include (i) the design of the reconstruction task, (ii) the design of the pseudo anomalies, and (iii) the design of the anomaly evaluation function.

\begin{table*}[]
\footnotesize
\caption{Comparison of image-level detection results on the MVTec-AD dataset when applying different pseudo anomalies}
\label{Table7}
\centering
\begin{tabular}{c|cccc|ccc}
\hline
Method     & \multicolumn{4}{c|}{EdgRec}                & \multicolumn{3}{c}{DeAE}       \\ \hline
\diagbox{Metric}{noise}     & DTD+Cutpaste & DTD  & Cutpaste & no\_noise & DTD+Cutpaste & DTD  & Cutpaste \\ \hline
AUC\_image & \textbf{97.8}         & 97.3 & 96.1     & 95.7      & 95.1         & 92.5 & 89.2     \\
AUC\_pixel & \textbf{97.7}         & 96.4 & 97.5     & 95.4      & 95.8         & 91.9 & 92.9     \\
AUPRO      & \textbf{92.1 }        & 91.1 & 91.1     & 87.8      & 89.1         & 83.6 & 80.2     \\
AP         & \textbf{65.8}         & 65.5 & 63.8     & 56.8      & 56.5         & 47.4 & 42.0     \\ \hline
\end{tabular}
\end{table*}

\begin{table}[]
\footnotesize
\centering
\caption{Comparison of image-level detection results on the MVTec-AD dataset when applying different anomaly evaluation functions}
\label{Table8}
\scalebox{0.9}{
\begin{tabular}{c|ccccc}
\hline
\diagbox{Metric}{function}     & $L_2$   & SSIM & MSGMS & color & MSGMS+color \\ \hline
AUC\_image & 93.3 & 89.7 & 96.9  & 83.1  & \textbf{97.8}        \\
AUC\_pixel & 96.1 & 94.5 & 97.3  & 86.1  & \textbf{97.7 }       \\
AUPRO      & 87.5 & 84.8 & 91.1  & 66.0  & \textbf{92.1  }      \\
AP         & 87.5 & 57.7 & 63.2  & 42.6  & \textbf{65.8 }       \\ \hline
\end{tabular}}
\end{table}

\subsection{The design of the reconstruction task}
\begin{figure}
    \centering
		\includegraphics[scale=0.25]{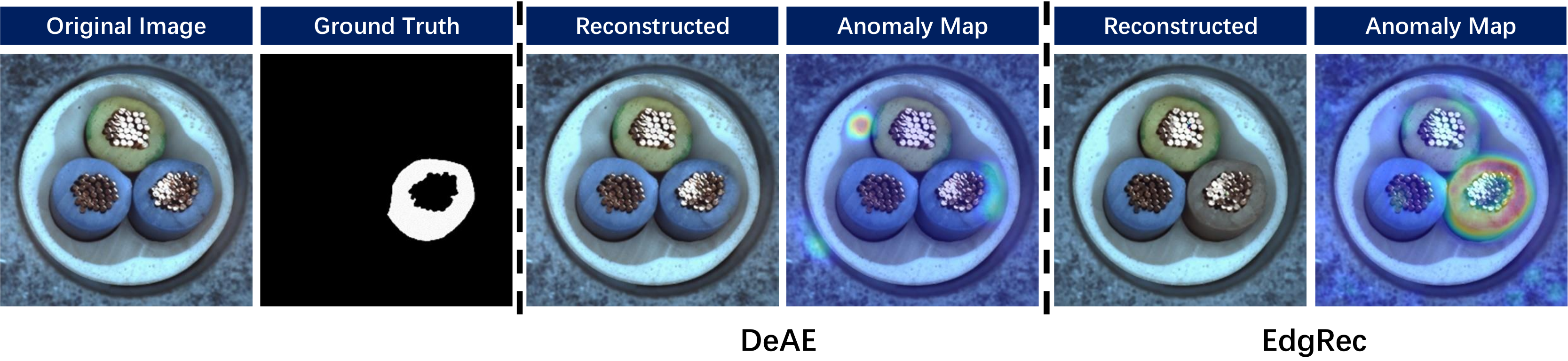}
	\caption{The comparison of qualitative results between the conventional denosing autoencoder DeAE and our EdgRec.}
	\label{FIG:4}
\end{figure}

Different from the conventional image-to-image reconstruction task, in this paper, we design a new edge-to-image restoration task. To verify its superiority over the conventional reconstruction network in anomaly detection, we conduct experiments named DeAE where we use the conventional image-to-image denoising autoencoder with all other settings unchanged. As can be seen in Table \ref{Table7}. (the column of the `DTD and Cutpaste'), our design improves the overall performance for all evaluation metrics. To visualize the specific difference in reconstructed images, we choose a typical example from the category cable, as can be seen in Fig. \ref{FIG:4}. Our network can better detect `cable swap' since we force the network to memorize the normal colors during the restoration, while the conventional network fails in detecting this type of defect. On the other hand, the conventional denoising autoencoder still achieves a comparable performance among existing baselines. This can mainly be attributed to the well-designed pseudo anomalies, which will be discussed in the following.

\subsection{The influence of the pseudo anomalies}
\begin{figure}
    \centering
		\includegraphics[scale=0.25]{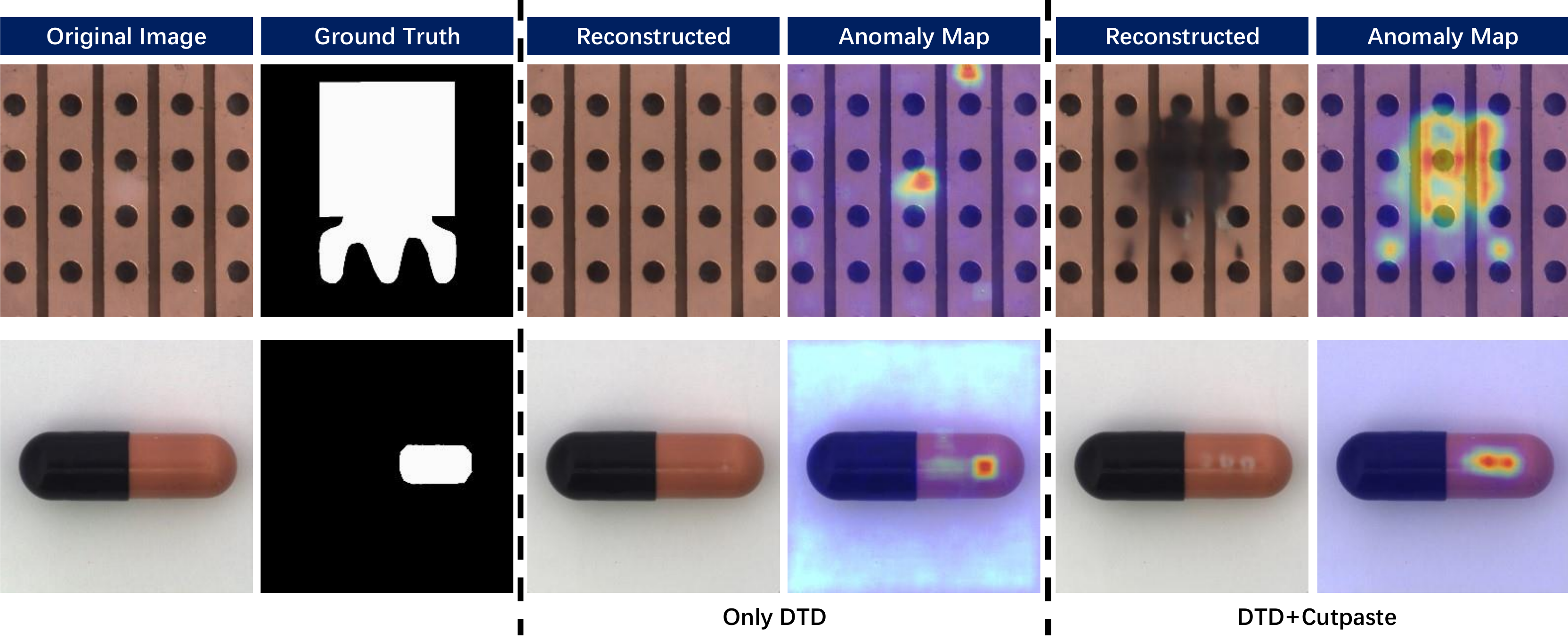}
	\caption{The comparison of qualitative results when applying different psuedo anomalies}
	\label{FIG:5}
\end{figure}

We propose to combine the DTD textures with Cutpaste augmentation to generate pseudo anomalies. To study their influence, we conduct experiments with different settings of pseudo anomalies for both EdgRec and DeAE. As can be seen in Table \ref{Table7}, the anomalies generated from DTD with the strategy proposed in \cite{zavrtanik2021reconstruction} can effectively improve the overall performance, especially for the conventional denoising autoencoder. When only using the Cutpaste augmentation, our EdgeRecon outperforms DeAE by a large margin. In particular, even without adding any pseudo anomalies, we can still achieve comparable results. These emphasize the contribution of our restoration task design. Also, the proposed CutPaste augmentation can further improve the performance, since it can help better detect the large global anomalies, as can be seen in Fig. \ref{FIG:5}.

\begin{figure}
    \centering
		\includegraphics[scale=0.3]{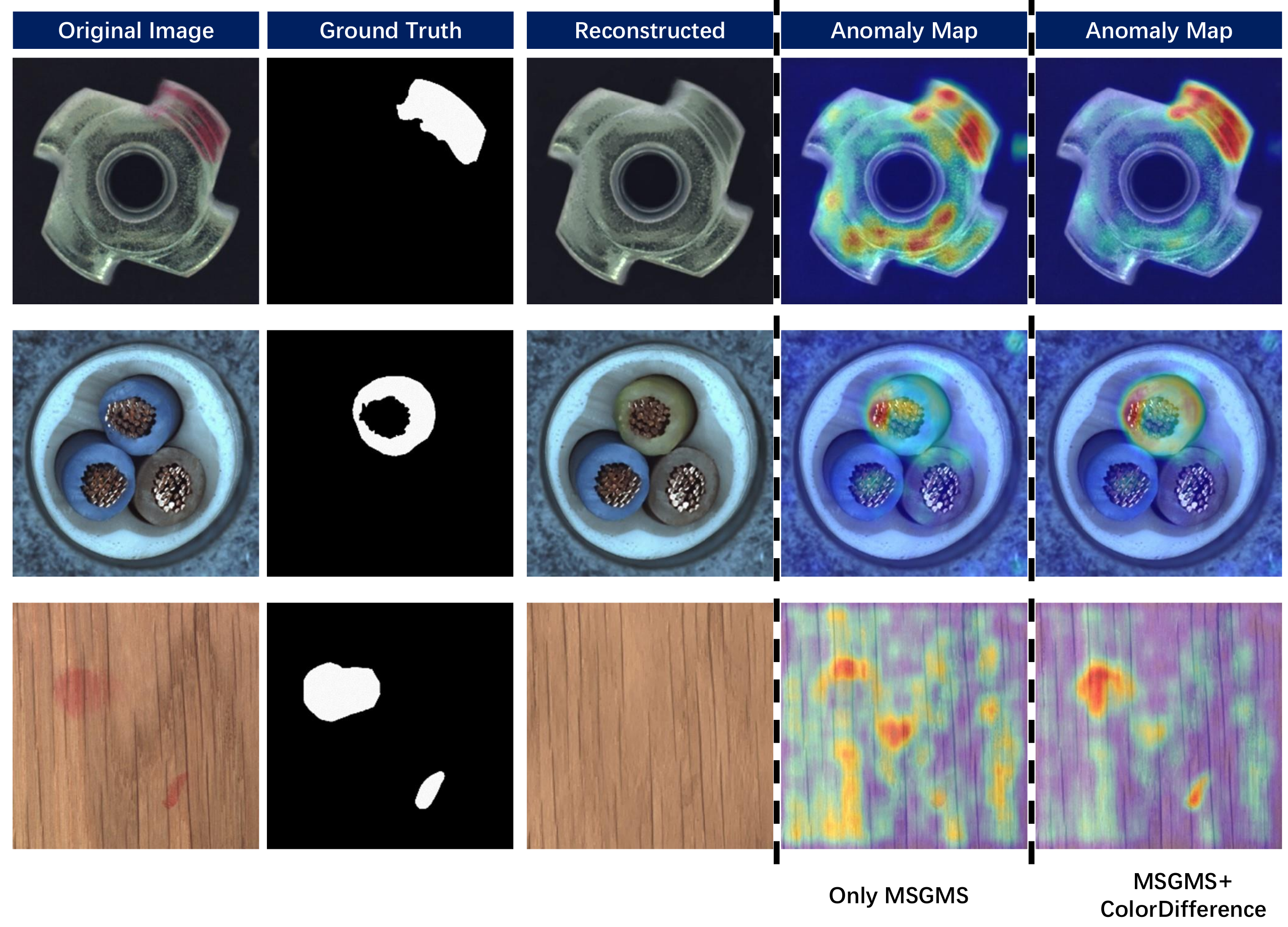}
	\caption{The comparison of qualitative results when applying different anomaly evaluation functions}
	\label{FIG:6}
\end{figure}

\subsection{The influence of the anomaly evaluation function}
We propose a new color-based anomaly evaluation function and combine it with the MSGMS to build our anomaly evaluation function. Here we conduct experiments to study the influence of different anomaly evaluation functions. As can be seen in Table \ref{Table8}, the combination of the MSGMS and our color-based function can significantly boost the overall performance. On the other hand, the color-based function cannot be used alone since many anomalies do not differ from normal regions in color. In practice, our anomaly evaluation function can be adjusted by reweighting different components (the gradient and the color) so as to adapt to different detection requirements. Fig. \ref{FIG:6} shows some visualized results under different anomaly evaluation functions.

\section{Conclusion}
This paper proposes a new reconstruction-based industrial anomaly detection model. Specifically, we design a new reconstruction task, i.e., reconstruct the RGB image from its grayscale edge. Besides, we design a new color-based anomaly evaluation function that can effectively detect color anomalies. The proposed method achieves convincing performance on the MVTec AD and MVTec 3D-AD datasets for both anomaly detection and localization. Compared to existing state-of-the-art methods, our method is more interpretable and tunable since we all use the hand-crafted anomaly evaluation function which can be easily adjusted for the specific industrial scenario.

We speculate that the performance of the reconstruction-based method can be further improved by introducing multi-scale image reconstruction techniques, more robust anomaly evaluation functions, and more diversified pseudo anomaly generation methods, which will be our future work.

\section*{Declaration of Competing Interest}
The authors declare that they have no known competing financial interests or personal relationships that could have appeared to influence the work reported in this paper.

\section*{Acknowledgement}
The calculations were performed by using the HPC Platform at Xi’an Jiaotong University. The article is fund by Large Instrument and Equipment Operation Subsidy Fund of Xi'an jiaotong University.




%\begin{thebibliography}{00}
%\bibliographystyle{elsarticle-num} 
%\bibliography{ref}

%\end{thebibliography}
\end{document}